\documentclass{article}
\usepackage{ijcai20}

\usepackage{times}
\usepackage{soul}
\usepackage{url}
\usepackage[hidelinks]{hyperref}
\usepackage[utf8]{inputenc}
\usepackage[small]{caption}
\usepackage{graphicx}
\usepackage{amsmath}
\usepackage{amsthm}
\usepackage{amssymb}
\usepackage{booktabs}
\usepackage{algorithm}
\usepackage{algorithmic}
\usepackage{subfigure}
\usepackage{xcolor}
\usepackage{graphicx}
\usepackage{multirow}

\newcommand{\N}{\mathcal{N}}

\newcommand{\ms}[1]{\boldsymbol{#1}}
\newcommand{\tabitem}{~~\llap{\textbullet}~~}

\title{Discrete Variational Attention Models for Language Generation}

\author{
  Xianghong Fang$^{1}$ \footnote{Equal contribution in random order.}\and
 Haoli Bai$^{1\hspace{0.7ex}*}$\and
 Zenglin Xu$^2$\and
 Michael Lyu$^{1}$\and
 Irwin King$^{1}$
 \affiliations
 $^1$The Chinese University of Hong Kong \\
 $^2$Harbin Institute of Technology, Shenzhen  \\
\emails
xianghong\_fang@163.com,
\{hlbai,lyu,king\}@cse.cuhk.edu.hk,
zenglin@gmail.com
 }

\begin{document}
\maketitle

\begin{abstract}
Variational autoencoders have been widely applied for natural language generation, however, there are two long-standing problems:  information underrepresentation and posterior collapse. The former arises from the fact that only the last hidden state from the encoder is transformed to the latent space, which is insufficient to summarize data. The latter comes as a result of the imbalanced scale between the reconstruction loss and the KL divergence in the objective function.
To tackle these issues, in this paper we propose the discrete variational attention model with categorical distribution over the attention
mechanism owing to the discrete nature in languages. Our approach is combined  with an auto-regressive prior to capture the sequential dependency from observations, which can enhance the latent space for language generation.
Moreover, thanks to the property of discreteness, the training of our proposed  approach does not suffer from posterior collapse.
Furthermore, we carefully analyze the superiority of discrete latent space over the continuous space with the common Gaussian distribution.
Extensive experiments on language generation demonstrate superior advantages of our proposed approach in comparison with the state-of-the-art counterparts.

\end{abstract}

\section{Introduction}


As one of the representative of deep generative models, variational autoencoders~(VAEs)~\cite{Kingma2013AutoEncodingVB} have been widely applied in natural language generation~\cite{Wang2019NeuralGC,Fu2019CyclicalAS,Li2019ASE}. Given input text $x$, VAE learns the variational posterior $q(z|x)$ through the encoder, and reconstructs the output from latent variables $z$ via the decoder $p(x|z)$. To generate diverse sentences, the decoder $p(x|z)$ heavily relies on the samples drawn from the prior $p(z)$ that controls the contents, topics and semantics for generation. However, despite the successful applications of VAEs for language generation, they majorly suffers from two long-standing challenges, i.e., information underrepresentation and posterior collapse. 


First of all, in most variational language generation methods~\cite{Fu2019CyclicalAS,He2019LaggingIN,Wang2019NeuralGC,Li2019ASE}, the latent space is derived from only the last hidden state of the encoder, and is therefore insufficient to summarize the input. We call this challenge as information underrepresentation.
Intuitively, given an observed sentence, the corresponding sequence of hidden states should be semantically correlated and representative during the phase of language generation. Thus a potential solution is to enhance the representation power
via the attention mechanism~\cite{Bahdanau2014NeuralMT}, which can build the correlation between the hidden states of the encoder and decoder. However, little efforts have been paid towards the utilization of attention in variational language generation. 

The next challenge is posterior collapse, a long standing phenomenon troubling the training of VAEs especially in the case of language generation. When the posterior fails to encode any knowledge from the observations,  the decoder therefore receives no signal at each time step during generation. 
Many approaches have been proposed to alleviate the issue, for instance, annealing the KL divergence term~\cite{Bowman2015GeneratingSF,Kingma2017ImprovedVI,Fu2019CyclicalAS}, revising the model~\cite{Yang2017ImprovedVA,Semeniuta2017AHC,Xu2018SphericalLS} and modifying the training procedure~\cite{He2019LaggingIN,Li2019ASE}. Despite the effectiveness of these methods, the trade-off between the reconstruction loss and the KL divergence is inevitable, and the phenomenon could still happen when the two terms are not properly scaled.


Aiming to address the above challenges, we propose the Discrete Variational Attention Model~(DVAM) with categorical distributions over the attention mechanism. 
As shown in Figure~\ref{fig:vam model}, the proposed DVAM
adopts two different RNNs as the encoder network and the decoder network, respectively. In order to better explore the prior context information for the language generation phase, we introduce the latent stochastic variable $z$ to build the semantic connection between the encoder hidden states and the decoder hidden states, via the attention~\cite{Bahdanau2014NeuralMT}. 
Considering that text inputs $x$ are more naturally modeled as a
sequence of discrete symbols rather than continuous ones, we explore the potential of quantized latent space in the attention mechanisum. The advantages of quantized representation have also been empirically justified in the recently proposed vector quantized variational autoencoder~(VQVAE)~\cite{Oord2017NeuralDR,Roy2018TheoryAE}  in learning the sequentially correlated prior for image generation.
We further show that the quantized representation can avoid DVAM trapping in posterior collapse (when  KL divergence $D_{KL}(q(z|x)||p(z))\rightarrow 0$)---since  the variational posterior $q(z|x)$ is a discrete distribution, the KL divergence is not differentiable with its parameters and hence not involved during model training.
This can also let us learn the variational posterior and prior separately. We first train DVAM until convergence where the posterior successfully encodes the sequential dependency from observations. Then we deploy informative context priors, such as a separate auto-regressive prior~\cite{Oord2016PixelRN}, to learn the sequential dependency from the well-trained posterior, after which we can sample diverse and representative latent sequences from the prior for sentence generation.
Furthermore, we provide detailed analysis on  the advantages of the quantized latent space over continual latent space. 
Finally, we evaluate the proposed model on several benchmark datasets for language modelling, and experimental results demonstrate the superiority of DVAM in generating languages over its counterparts.

Our contributions can thus be summarised as:
\begin{enumerate}
	\item We propose the discrete variational attention model with an auto-regressive prior to capture the sequential dependency in the latent space, such that issues of information underrepresentation and posterior collapse can be effectively tackled.
	\item We carefully analyze reasons why discrete latent space with categorical distribution in our model is preferred than the  continuous space with the commonly Gaussain distribution for language generation.
	\item Experimental results on benchmark datasets demonstrate the advantages of our DVAM against state-of-the-art baselines in language generation.
\end{enumerate}


\section{Background}
\subsection{Variational Antoencoders for Language Generation}
Variational Autoencoders~(VAEs)~\cite{Kingma2013AutoEncodingVB} are well known class of generative models. Given observations $x$, we seek to infer latent variables $z$ from which new observations $\hat{x}$ can be generated. To achieve this, we need to maximize the marginal log likelihood $\log p_\theta(x)$, which is usually intractable due to the complex posterior $p(z|x)$. Consequently an approximate posterior $q_{\phi}(z|x)$ (i.e. the \textit{encoder}) is introduced, and the evidence lower bound (ELBO) of the marginal likelihood is maximized as follows:
\begin{equation}
\log p_\theta (x) \geq \underbrace{\mathbb{E}_{z \sim q_{\phi}(z|x)}[\log  p_{\theta}(x|z)]}_\textrm{reconstruction loss} - \underbrace{ D_{KL}(q_{\phi}(z|x) \Vert p(z))}_\textrm{KL divergence},
\label{eq:elbo}
\end{equation}
where $p_{\theta}(x|z)$ represents likelihood function conditioned on latent codes $z$, also known as the \textit{decoder}. $\theta$ and $\phi$ are the corresponding parameters.

In language generation, VAEs are used to learn the mapping from the latent space of $z$ to the observations $x$. Based on this mapping, new sentences can be effectively generated. 
VAEs are usually armed with an RNN encoder and decoder, where the input sentences $x$ are given  to the encoder, the latent variables $z$ are derived from the last hidden states $h^e_T$ via the reparameterization trick~\cite{Kingma2013AutoEncodingVB} $z = \mu(h^e_T) + \sigma(h^e_T)\epsilon$ with $\epsilon\sim\N(0,I)$. To generate new sentences, we sample latent variables $z$ from the prior that are forwarded to the decoder to generate sequences $\hat{x}$ by
\begin{equation}
p(\hat{x}_{1:T}| z) = \prod_{t=2}^{T} p(\hat{x}_{t}|\hat{x}_{t-1}, z)\cdot p(\hat{x}_1|z).
\label{eq:decoder_generation}
\end{equation}

\subsection{ Diminishing Effect of Latent Variables}
Modeling RNN-based language models with VAEs is prone to suffer from the \textit{diminishing effect} of latent variables, i.e., the generated sequences $\hat{x}$ are loosely dependent on $z$ and thereon the learned latent space plays no role in generation. The diminishing effect largely comes from two aspects:

\paragraph{Information Underrepresentation.} 
While sentences $x$ are fed into the encoder, the corresponding latent variables $z$ are obtained through the transformation of only last hidden state $h^e_T$ of the encoder. The resulting latent space, however, is usually insufficient to summarize the observations $x$. The rich semantics in the whole sequences are thereon lost in $z$, known as the information underrepresentation.
During generation, such latent variables $z$ are forwarded to the decoder, which cannot effectively guide the decoder to generate sentences with high correlation and quality.
Some recent attempts~\cite{bahuleyan2017variational,deng2018latent} borrow ideas from the attention mechanism~\cite{Bahdanau2014NeuralMT}, and introduce variational distributions over the context vector, however, they use uninformative prior that may fail to sample sequentially correlated latent variables for sentence generation. 



\paragraph{Posterior Collapse.}
Posterior collapse usually arises as $D_{KL}(q_{\phi}(z|x) \Vert p(z))$ in Equation~(\ref{eq:elbo}) diminishes to zero, where the local optimal gives $q_{\phi}(z|x) = p(z)$. When posterior collapse happens, we can verify that $x$ are independent of $z$ by $p(x)p(z) = p(x)q_{\phi}(z|x) = p(x)\frac{p(x, z)}{p(x)} = p(x, z)$. Therefore, the encoder learns a data-agnostic posterior (i.e., the standard Normal distribution) without any information from the observations $x$, while the decoder learns to generate itself without actually relying on the latent variable $z$.

Posterior collapse happens inevitably as the ELBO contains both the reconstruction loss $\mathbb{E}_{z \sim q_{\phi}(z|x)}[\log  p_{\theta}(x|z)]$ and the KL-divergence $D_{KL}(q_{\phi}(z|x) \Vert p(z))$. When the scales between the two terms are not properly balanced, it could easily make the KL divergence over optimized. Moreover, given a powerful and auto-regressive decoder, the decoder itself can learn to generate sentences $\hat{x}$ without actually relying on $z$.


\section{Methods}
\begin{figure*}
    \vspace{-3ex}
	\centering
	\includegraphics[width=0.9\textwidth]{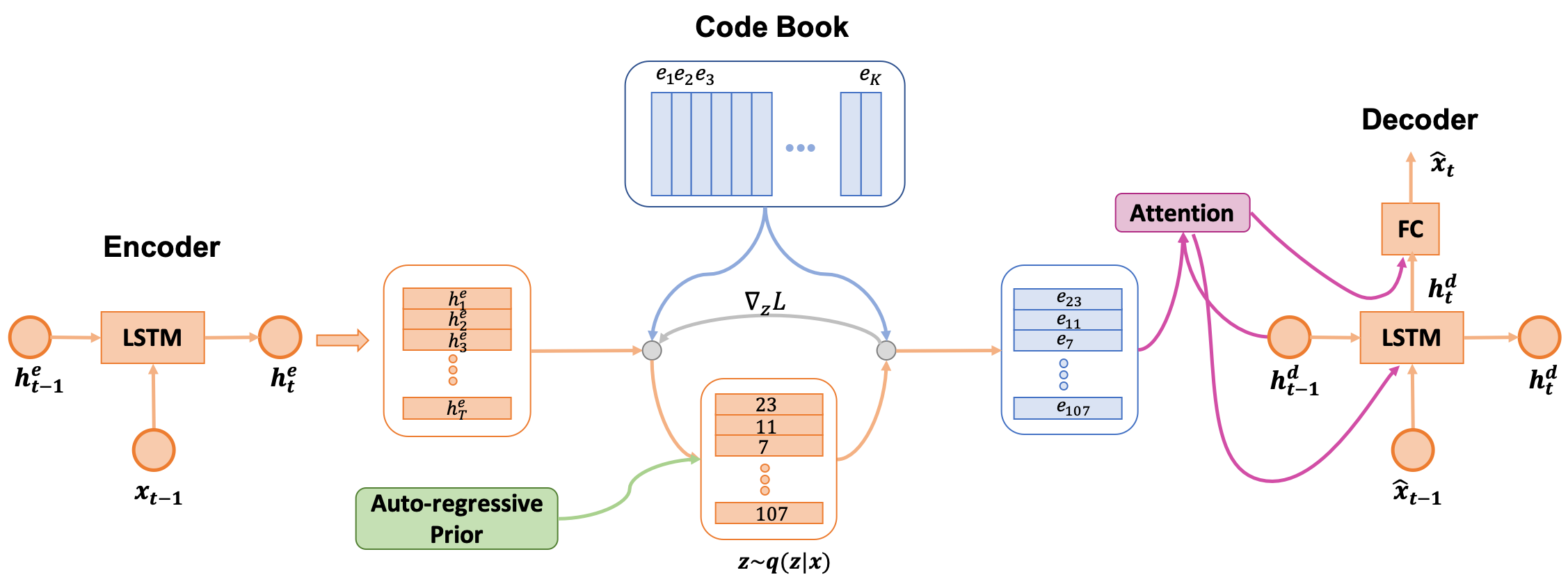}
	\vspace{-2ex}
	\caption{The overall architecture of the proposed DVAM. Given observations $x$, the encoder first maps $x$ to hidden states $h^e_{1:T}$, which are then quantized based on Euclidean distance to the code book. The quantized hidden states $e_{z_{1:T}}$ are then forwarded to the attention module to align the decoder.
	To generate new sentences from DVAM, we deploy an auto-regressive prior to draw sequentially correlated samples.}
	\label{fig:vam model}
	\vspace{-3ex}
\end{figure*}


In order to tackle the diminishing effect of latent variables in variational language models, we first present the Discrete Variational Attention Model~(DVAM), and then introduce meaningful priors for language generation. 

\subsection{Discrete Variational Attention Models}
\label{sec:vam}

As shown in Figure~\ref{fig:vam model}, the proposed DVAM
adopts two different RNNs as the encoder network and the decoder network, respectively. In order to build the connection between the encoder hidden states (denoted by $ h^e_{1:T}$) and the decoder hidden states (denoted by$ h^d_{1:T}$), we further involve an attention mechanism similar to sequence to sequence~(seq2seq) models~\cite{Bahdanau2014NeuralMT}. 



A proper variational posterior $q_\phi(z_{1:T}|x)$ plays an important role in effectively capturing the semantic information from the observations $x$. A simple idea would be choosing the widely used Gaussian distribution, which however easily leads to posterior collapse (as will be discussed in Section~\ref{sec:advantages}). As representations of languages are discrete in nature, we seek to
quantize the latent space with a code book $\{e_k\}_{k=1}^K$, where
$K$ is the code book size. The combination of code book vectors therefore represents the sequential dependency from observed sentences x.
The advantages of quantized representation have also been verified in the recently proposed vector quantized variational autoencoder~\cite{Oord2017NeuralDR,Razavi2019GeneratingDH} for image generation.
We set the variational posterior  $q_\phi(z_{1:T}|x)$
as categorical distribution to index $\{e_k\}_{k=1}^K$ as follows:

\begin{equation}
\label{eq:vq_distribution}
q_\phi(z_t=k| x) =
\begin{cases}
1 & \text{for } k=\arg\min_{j}\|h^e_t-e_j\|_2\\
0 & \text{otherwise}
\end{cases}.
\end{equation}

Given the code book index $z_t$, the encoder hidden state $h_t^e$ is therefore quantized to $e_{z_t}$, and the attention scores can be computed as usual by 
\begin{equation}
\hspace{2ex} \alpha_{ti} = \frac{\exp(\tilde{\alpha}_{ti})}{\sum_{j=1}^{T}\exp(\tilde{\alpha}_{tj})},
\end{equation}
where $\tilde{\alpha}_{t}=v^{\top}\tanh(W_e e_{z_{1:T}} + W_d h^d_{t-1} + b)$ is the score before the softmax normalization. 

We then compute the context vectors $c_t = \sum_{i=1}^{T} \alpha_{ti} e_{z_i}$ as an extra input to the decoder, and reformulate the generation process as

\begin{equation}
p(\hat{{x}}_{1:T}| z_{1:T}) = \prod_{t=2}^{T} p(\hat{x}_{t}|\hat{x}_{t-1}, z_{1:T})p(\hat{x}_1| z_{1:T}).
\label{eq:decoder_generation}
\vspace{-0.5ex}
\end{equation}
Therefore at each time step, the decoder receives the supervision from the context vector $c_t$ , which is a weighted sum from the code books $\{e\}_{k=1}^K$ of the whole sequence.
Consequently, the variational posterior  $q_\phi(z_{1:T}|x)$ encodes the sequential dependency from the observations that addresses the issue of information underrepresentation.

To allow new sentence generation, our model can also draw sequentially dependent samples from the prior $p(z_{1:T})$, which will be discussed in the next section.


\subsection{Model Training}
Thanks to the nice property of discreteness, the optimization of DVAM does not suffer from posterior collapse. 
To see this, note that the ELBO in Equation~(\ref{eq:elbo}) includes both the reconstruction loss and the KL divergence, whereas the KL divergence of DVAM can be written as 
\begin{align}
    \label{eq:vqvae_kl}
    & D_{KL}(q_\phi( z_{1:T}| x)||p( z_{1:T})) \nonumber\\
    &= \sum_{t=1}^T \{ - H(q_\phi( z_t)) - \sum_{k=1}^K q_\phi( z_t=k|  x)\log p(z_t=k) \} \nonumber\\
    &= - 0 - \sum_{t=1}^T \{ 1\cdot \log p( z_t=j_t) + 0 \cdot \log p(z_t\neq j_t)\}, \
\end{align}
where the third line is obtained by $q_\phi( z_t=j_t)=1$, $q_\phi(z_t\neq j_t)=0$ and the fact that $H(q_\phi( z_t)) = -1 \log 1 - 0 \log 0 = 0$. Consequently, $D_{KL}(q_{\phi}(z|x) \Vert p(z))$ is not differentiable w.r.t. the variational parameters $\phi$. The optimization of KL divergence does not play a role in the reconstruction loss optimization, and even we do not need to know the form of the prior $p(z_{1:T})$ in advance.

On the other hand, since the latent variables $z_{1:T}$ are obtained based on the Euclidean distance to the code books, we encourage the hidden states to stay close to $\{e_k\}_{k=1}^K$. Therefore, the training objective for DVAM can be formulated as
\begin{equation}
    \label{eq:vqvae_loss}
    \min_{\theta, \phi} -\mathbb{E}_{ z_{1:T}\sim q_{\phi}} 
    \log p_{\theta}( x| z_{1:T}) + \beta\sum_{t=1}^{T}\| h_t^e - \mathrm{{sg}}(e)\|_F^2,
\end{equation}
where $\beta$ is the regularizer, and $\mathrm{{sg(\cdot)}}$ stands for stop-gradient operation.
Note that since the quantization is non-differentiable, we adopt the widely used straight through estimator~(STE)~\cite{bengio2013estimating}] to copy gradients from $e_{z_t}$ to $h_{t}^{e}$, as is shown in Figure~\ref{fig:vam model}, 
In terms of the code books $\{e_k\}_{k=1}^K$, since Equation~(\ref{eq:vqvae_loss}) is non-differentiable with them, we first apply the K-means algorithm to calculate the average over all latent variables $h_{1:T}^e$ that are closest to $\{e_k\}_{k=1}^K$, then we take exponential moving average over the code books so as to stabilize the mini-batch update.

\paragraph{Auto-regressive Prior for Language Generation} 
Despite the optimization of Equation~(\ref{eq:vqvae_loss}) does not rely on the choice of the prior, however, the form of prior does affect the process of language generation.
Recall that in the training phase, latent variables $ z_{1:T}\sim q_\phi( z_{1:T}| x)$ are sampled conditioned on the input $x$, and therefore the posterior learns to encode the sequential dependency in the latent space.
Nevertheless, to generate new sentences after network training, we first sample $z_{1:T}$ unconditionally from the prior and then pass it to the decoder for generation, where the sequential dependency can hardly be guaranteed. As a result, the decoder cannot receive the structured supervision from the latent space for valid generation.

To solve the problem, we seek to find a auto-regressive prior $p_\psi(z_{1:T}) = p_\psi(z_1)\prod_{t=2}^T p_\psi(z_t|z_{1:t-1})$ parameterized by $\psi$ such that it has enough capacity to capture the underlying sequential structures in the posterior $q_\phi(z_{1:T}|x)$.
Towards that end, we adopt a PixelCNN~\cite{Oord2016ConditionalIG} to learn the prior. Unlike PixelCNN models on images, we use a 16-layer residual 1-dimensional convolutional network to swap over the latent sequence. 
In order to learn the sequential dependency in the posterior $q_\phi(z_{1:T}|x)$, we first train DVAM using Equation~(\ref{eq:vqvae_loss}) until convergence, and then minimize the KL divergence $\sum_{t} D_{KL}(q_\phi( z_t|  x)||p_\psi( z_t|z_{1:t-1}))$ w.r.t $\psi$, which reduces to the cross entropy loss according to Equation~(\ref{eq:vqvae_kl}).

\subsection{Discussion on the Distribution of $z$}
\label{sec:advantages}
As mentioned earlier in Section~\ref{sec:vam}, a simple idea is to directly assign Gaussian distributions over the attention, i.e. $ z_t = \mu_t(h^e_t) + \sigma_t(h^e_t) \epsilon$ for  $\epsilon\sim\N(0, I)$, leading to the Gaussian Variational Attention Model~(GVAM). 

To analyze the defects of GVAM, we first
recall the KL divergence between two Gaussian distributions, which can be written as follows:
\begin{align}
\label{eq:gaussian_kl}
& \sum_{t=1}^{T}D_{KL}(q_{\phi}({z_t}|{x}) \Vert p_\psi({z_t|z_{1:t-1}})) \nonumber\\
& = \sum_{t=1}^T \sum_{d=1}^D \frac{1}{2} (\log \frac{\hat{\sigma}_{td}^2}{\sigma_{td}^2} - 1 + \frac{\sigma_{td}^2 + (\hat{\mu}_{td} - \mu_{td})^2}{\hat{\sigma}_{td}^2}),
\end{align}
where $D$ is the latent dimension of $z_t$. We denote $\{\mu_{td}, \sigma_{td}\}$ and $\{\hat{\mu}_{td}, \hat{\sigma}_{td}\}$ as the parameters of the posterior and prior distributions, respectively.
Unlike the objective of DVAM in Equation~(\ref{eq:vqvae_kl}), we find that Equation~(\ref{eq:gaussian_kl}) is differentiable w.r.t to variational parameters $\phi$, and therefore should be included for the minmization of ELBO.
As we also need an auto-regressive prior during generation, such differentiation brings the challenge for GVAM.
Unlike DVAM that learns the posterior $q_\phi(z_{1:T}|x)$ in advance to teach the prior, GVAM needs to involve the posterior and prior jointly during the optimization, i.e., 
\begin{equation}
    \min_{\phi,\theta,\psi} -\mathbb{E}_{ z_{1:T}\sim q_{\phi}} 
    \log p_{\theta}( x| z_{1:T}) + D_{KL}(q_{\phi}(z_{1:T}|{x}) \Vert p_\psi({z_{1:T}})).
\end{equation}

Such optimization can be easily troubled by posterior collapse due to two aspects. Firstly, the scale of the KL divergence increases linearly to the length of the sequence, and a large scale usually overlooks the minimization of the reconstruction loss. The training thereon is unstable across various observations $x$ with different lengths.
Secondly and more seriously, both $\phi$ and $\psi$ are used to minimize the KL divergence. Whenever $q_\phi(z_{1:T}|x)$ collapses to $p_\psi(z_{1:T})$ before the $q_\phi(z_{1:T}|x)$ learns any sequential dependency from the observations, both $q_\phi(z_{1:T}|x)$ and $p_\psi(z_{1:T})$ are trapped to the local optimal that cannot provide structural supervision to the decoder during both training and generation.

\section{Experiments}

In this section, we verify the advantages of our proposed DVAM for language generation.
We first perform language modelling on three benchmark datasets, which measures the model capacity of different approaches.
Then we investigate the training of the auto-regressive prior, and evaluate the generated sentences from these approaches.
Finally we conduct a set of ablation studies to shed more lights into DVAM. 
Codes implemented by Pytorch will be released on Github. 

\subsection{Experimental Setup}
We take three benchmark datasets of language modelling for verification: Yahoo Answers ~\cite{Xu2018SphericalLS}, Penn Tree~\cite{Marcus1993BuildingAL}, and a down-sampled version of SNLI~\cite{bowman2015large}. A summary of dataset statistics is shown in Table~\ref{table:datasets}.

\begin{table}[h]
\vspace{-2ex}
\caption{Dataset statistics.}
\footnotesize
\vspace{-2ex}
\begin{tabular}{c|cccc}
\hline
Datasets &Train Size&Val Size &Test Size & Avg Len\\\hline
Yahoo & 100,000 & 10,000 & 10,000& 78.7\\
PTB & 42,068 & 3,370 & 3,761  & 23.1 \\
SNLI & 100,000 & 10,000 & 10,000  & 9.7\\
\hline
\end{tabular}
\label{table:datasets}
\vspace{-3ex}
\end{table}


\begin{table*}[t]
\vspace{-1ex}
\begin{center}
\caption{Results of language modelling on Yahoo, PTB and SNLI Datasets.}
\vspace{-1ex}
\label{table:density_results}
\begin{tabular}{l|c|c|c|c|c|c|c|c|c}
\hline
\multirow{2}{*}{Method} & \multicolumn{3}{|c}{Yahoo} 
& \multicolumn{3}{|c}{PTB} &\multicolumn{3}{|c}{SNLI}\\
\cline{2-10}
& Rec$\downarrow$ & PPL$\downarrow$ & KL & Rec$\downarrow$ & PPL$\downarrow$ & KL & Rec$\downarrow$ & PPL$\downarrow$ & KL\\\hline
LSTM-LM & - & 60.75  & -  & - & 100.47  & - & - & 21.44  & - \\
VAE & 329.08 & 61.60 & 0.00 &  102.31 & 106.59 & 0.00 & 33.36 & 22.22 & 0.007\\
+anneal & 328.62 & 60.34 & 0.00  & 102.18  & 105.87 & 0.00 & 32.33 & 20.19 & 1.05\\
+cyclic & 328.78 & 61.35 & 0.06 & 102.07 & 105.36 & 0.00 & 31.71 & 19.07 & 1.65\\
+aggressive & 323.08  & 57.12 & 4.94 & 100.81 & 96.46 & 1.85 & 32.91 & 21.31 & 0.47\\
+FBP  & 326.58 & 59.68 & 8.08 & 98.06 & 89.91 & 5.91 & 31.91 & 19.41 & 1.99\\
+pretraining+FBP & 316.17  & 52.39 & 16.17 & 94.88 & 76.21 & 7.47 & 30.62 & 17.22 & 2.83\\\hline
GVAM & 350.14 & 79.28 & 0.00 & 102.20 & 105.94 & 0.00 & 30.90 & 17.68 & 0.38    \\
DVAM (K=128) & 303.65 & 44.36 & 1.88 & 79.94 & 38.38 & 2.21 & 16.08 & 4.46 & 2.33  \\
DVAM (K=512) & \textbf{259.68} & \textbf{25.83} & 2.60 & \textbf{64.79} & \textbf{19.22} & 3.13& \textbf{11.06} & \textbf{2.82} & 2.58  \\\hline
\end{tabular}
\end{center}
\vspace{-3ex}
\end{table*}



We compare the proposed DVAM against a number of baselines, including the classical 
LSTM-LM, vanilla VAE~\cite{Kingma2013AutoEncodingVB},  as well as its advanced variants, e.g. annealing VAE~\cite{Bowman2015GeneratingSF}, cyclic annealing VAE\footnote{\url{https://github.com/haofuml/cyclical_annealing}}~\cite{Fu2019CyclicalAS}, lagging  
VAE\footnote{\url{https://github.com/jxhe/vae-lagging-encoder}}~\cite{He2019LaggingIN}, Free Bits (FB)~\cite{Kingma2017ImprovedVI} and pretraining+FBP VAE\footnote{\url{https://github.com/bohanli/vae-pretraining-encoder}}~\cite{Li2019ASE}. 
We also compare to our closest counterpart, i.e.,  Gaussian Variational Attention Model~(GVAM)  so as to directly verify the advantages of discreteness in the variational attention model.


We evaluate the performance of language generation models using three metrics, i.e., the reconstruction loss (Rec) calculated as $\mathbb{E}_{\ms z \sim q_{\phi}(\ms{z} | \ms x)}[\log  p_{\theta}(\ms {x}|\ms{z})]$ for measuring the ability to recover data from latent space (the lower the better), the Perplexity (PPL) measuring the capacity of language modelling (the lower the better), and the KL divergence (KL) between the posterior $q(z|x)$ and the prior $p(z)$ indicating whether posterior collapse occurs. 

\subsubsection{Implementation}
For baselines, we keep the same hyper-parameter settings to pretraining+FBP VAE~\cite{Li2019ASE}, e.g., the latent dimension of $z$, the word embedding size as well as the hidden size of the LSTM.
Since our latent variables are discrete, we do not use importance weighted samples to approximate the reconstruction loss in Lagging VAE~\cite{He2019LaggingIN} and pretraining+FBP VAE~\cite{Li2019ASE}. Also, for all methods, we compute PPL by using the reconstruction loss instead of ELBO. Therefore we reproduce their results based on the released codes, which is slightly different from those in original papers. 
For GVAM and DVAM, we keep shared hyper-parameters of baselines unchanged. By default we set the code-book size $K$ to $512$.
We first warm up the training for 30 epochs, and then gradually increase $\beta$ in Equation~(\ref{eq:vqvae_loss}) from 0.1 to $\beta_{max} = 5.0$, in a similar spirit to annealing VAE. 
For all experiments, we use the SGD optimizer with learning rate $1.0$, and decay it until five counts if the loss on the validation set does not decrease for $2$ epochs. For the auto-regressive prior, we use the same architecture and training settings for both GVAM and DVAM.

\subsection{Results}

\subsubsection{Language Modelling}
We first perform language modelling over the testing corpus of benchmark datasets, which measures the capacity of different approaches. 
Generally, the model is more expressive when it achieves lower Rec and PPL on the observations.
We average the KL divergence for GVAM and DVAM along the sequence to make them comparable to rest baselines.
Note that unlike other approaches, the KL divergence of DVAM does not affect language modelling and the model capacity, but is only related to the training of the auto-regressive prior.

The results of language modelling on Yahoo, PTB and SNLI are listed in Table~\ref{table:density_results}.
Comparing to baselines without variational attentions, we find that despite the variational attention is adopted, GVAM does not show advantages over other baselines in Yahoo and PTB  datasets. 
The KL divergence of GVAM is as tiny as that of VAE, especially on Yahoo and PTB when the average sequence length is long. This indicates posterior collapse which fails to learn the sequential dependency in the observation. 
On the other hand, our DVAM achieves significantly better results on all three datasets, especially with large code book size $K$. The success verifies the fact that the variational posterior learns to adapt itself to the sequential dependency, which significantly improves the expressiveness for language modelling.

\begin{table*}[t]
\vspace{-1ex}
\begin{center}
\caption{Sampled Sentences on Yahoo Dataset.}
\vspace{-1ex}
\label{tab:generated_sentences}
\footnotesize
\begin{tabular}{l|l}
\toprule
Method & Samples \\\hline
pretraining+ &  
\tabitem i hate wandering, i just wan na know when the skies in the sky and the winds. [/s]\\
FBP & \tabitem where is it that morning when snow on thanksgiving ? what's the next weekend ? dress it!!!! my \\
VAE& mother was the teen mom and i love her and she just is going to be my show. [/s]\\ 
&\tabitem are they allowed to join (francisco) in \_UNK. giants in the first place.? check out other answers. \\
& do you miss the economy and not taking risks in the merchant form, what would you tell? go to \\
&the yahoo home page and ask what restaurants follow this one. [/s]\\
\hline
GVAM 
& \tabitem didn't i still worry, he loves porn and feels awful??? [/s]\\
& \tabitem if i aint divorced b4 the prom, and i wont worry, worry, i realy worry, and nobody feels awful, \\
& and i realy \_UNK sometime, i wont worry, and eventually. [/s]\\
& \tabitem what is the worst and worst moment and the worst and worst, and a stranger and hugs, and nobody,\\
&and i deserve it, and nobody feels awful and i wont worry and worry.  [/s]\\
\hline
DVAM & \tabitem i need to start a modeling company ! any suggestions on what is a reliable topic? [/s]\\
(K=512)& \tabitem does anyone agree, there is a global warming of the earth? in general. there are several billion things,\\
& including the earth, solar system. [/s]\\
& \tabitem is anyone willing to donate plasma if you are allergic to cancer or anything else? probably you can.\\
&i've never done any thing but it is only that dangerous to kill bacteria. i have heard that it doesn't have \\
&any effect on your immune system. [/s]\\
\bottomrule
\end{tabular}
\end{center}
\vspace{-2ex}
\end{table*}

\subsubsection{Training Dynamics}
\begin{figure}[h]
    \vspace{-1ex}
	\centering
	\includegraphics[width=0.42\textwidth]{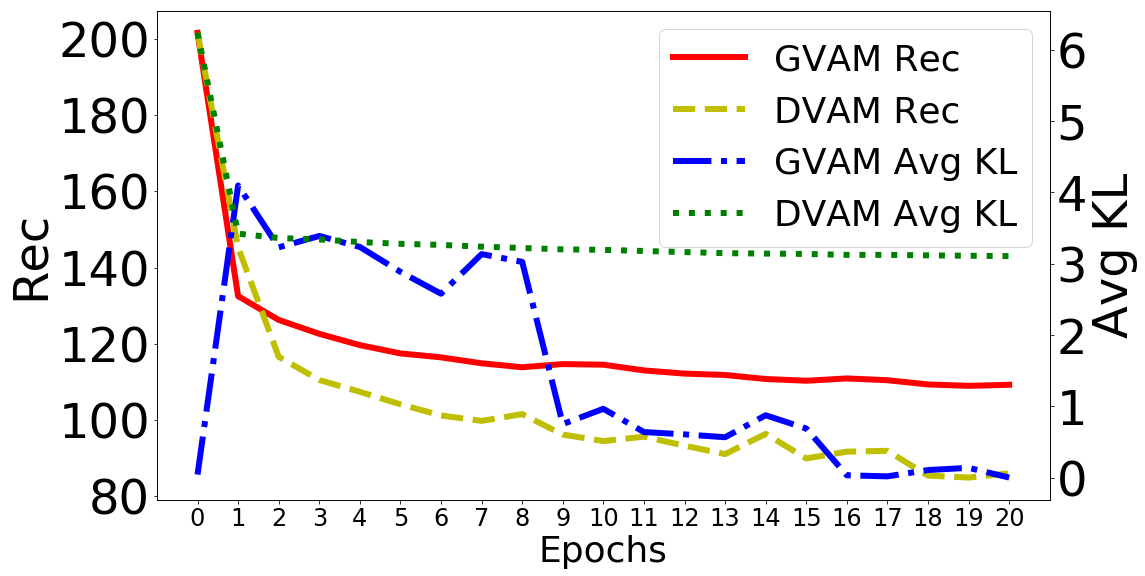}
	\vspace{-2ex}
	\caption{The loss curvature during first 20 epochs on PTB.}
	\label{fig:dvam_gvam_training}
	\vspace{-3ex}
\end{figure}

To further demonstrate the advantages of DVAM over GVAM, we now turn to investigate their training dynamics. We plot the  curvature of Rec and KL on the validation set of PTB in Figure~\ref{fig:dvam_gvam_training}.
We can find that the KL of GVAM tends oscillate at the beginning and diminishes quickly, whereas Rec does not decrease sufficiently. For DVAM, since Rec is not affected by the KL, Rec is minimized sufficiently. Meanwhile, the KL of DVAM converges quickly without oscillation as only parameters in the prior are updated to learn the sequential structure in the posterior.



\begin{figure*}[t]
    \vspace{-2ex}
	\centering
	\label{fig:ablation}
	\subfigure[Code book size $K$] { 
		\label{fig:two_stage}     
		\includegraphics[width=0.3\textwidth]{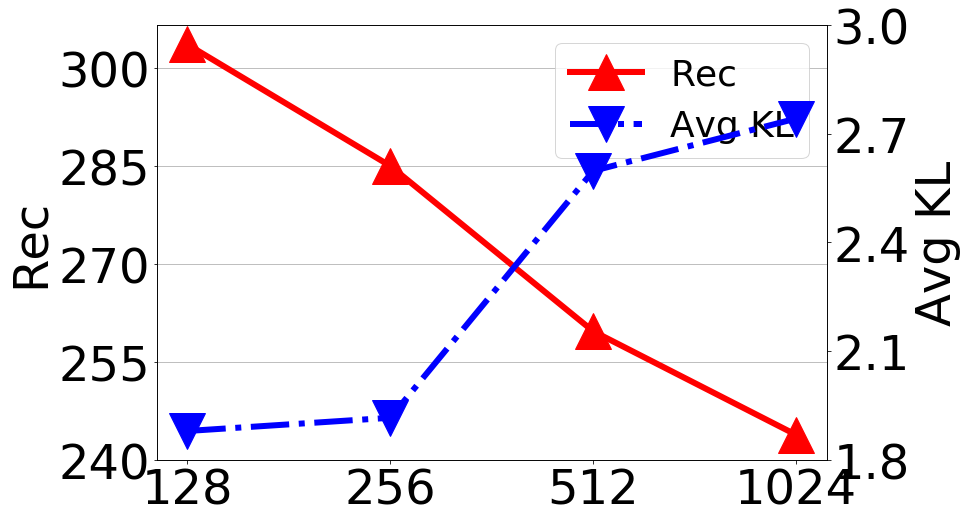}     
	}
	\subfigure[Maximum regularizer $\beta_{\max}$]{
		\label{fig:vq_weight}
		\includegraphics[width=0.3\textwidth]{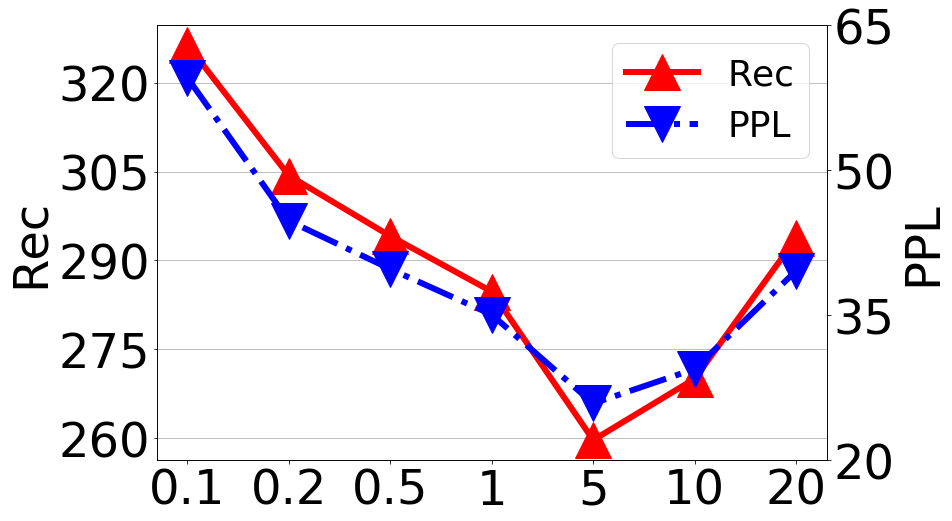}  
	}
	\subfigure[Latent dimension of $e_{z_t}$ ]{
		\label{fig:latent_dimension}
		\includegraphics[width=0.3\textwidth]{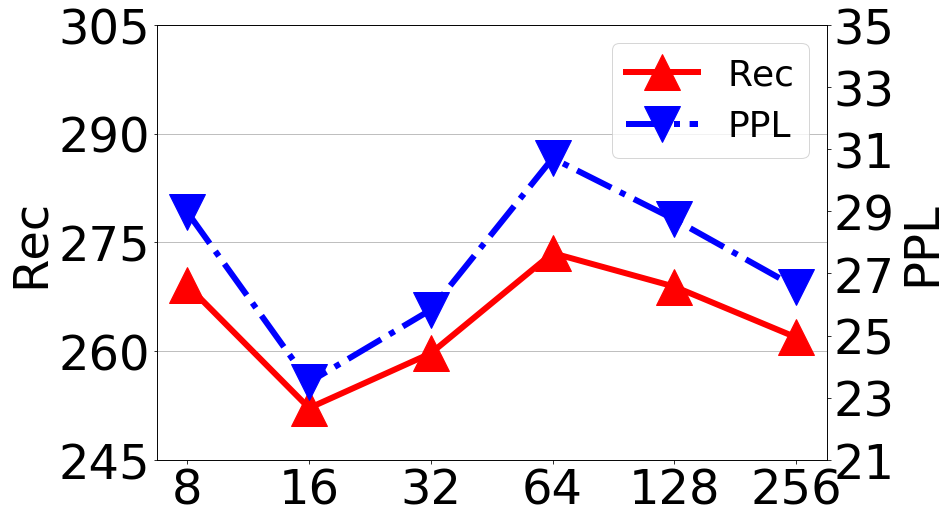}  
	}
	\vspace{-2ex}
	\caption{Ablation studies of DVAM.}
	\vspace{-2ex}
\end{figure*}

\subsubsection{Sampled Sentences}
Finally, we compare the generated sentences from the pretraining+FBP VAE, GVAM and our DVAM (K=512) respectively. Due to the limited space, we randomly sample 3 generated sentences with different lengths based on the models trained on Yahoo dataset, as is shown in Table~\ref{tab:generated_sentences}. We can find that while pretraining+FBP VAE produces readable sentences, the semantic meanings are poorly consistent. 
For GVAM, we can hardly find readable sentences even on short sequences, which is probably due to poorly correlated samples from the prior when both the posterior and prior are trapped to some local minimal.
On the contrary, our DVAM can produce relatively good sentences with consistent semantic meanings, even when the sequences are long. This suggests the samples from the prior indeed contain sequential dependency that benefits the generation.




\subsection{Ablation Studies}
By default, all the ablation studies are conducted on Yahoo dataset with the default parameter settings.


\subsubsection{Code Book Size $K$}
We begin with the effect of different code book size $K$ on the reconstruction loss and KL divergence for language modelling. We vary $K\in\{128,256,512,1024\}$, and the results are shown in Figure~\ref{fig:two_stage}. It can be observed that as $K$ increases, the Rec loss decreases while KL increases, both monotonically. The results are also consistent to Table~\ref{table:density_results} by increasing $K$ from 128 to 512.
Such phenomenons are intuitive, since a larger $K$ improves model capacity but
poses more challenges for training the auto-regressive prior.
Consequently, one should properly choose the code book size, such that the prior can approximate the posterior well, and yet the posterior is representative enough for the sequential dependency.




\subsubsection{Maximum Regularizer $\beta_{max}$}
Then we tune the maximum regularizer $\beta_{max}$, which controls the distance of the continuous hidden state $h_{1:T}^e$ to the code books $\{e_k\}_{k=1}^K$. Recall that a small $\beta_{max}$ loosely restricts the continuous space $h_{1:T}^e$ to the code book, making the quantization process hard to converge. On the other hand, if $\beta_{max}$ is too large, $h_{1:T}^e$ could easily get stuck in some local minimal during the training. Therefore, it is necessary to find a proper trade-off between the two situations. We vary $\beta_{max} \in \{0.1, 0.2, 0.5, 1,5,10, 20\}$, and the results is shown in Figure~\ref{fig:vq_weight}.
We can find that when $\beta_{max}=5$, the algorithm achieves the lowest Rec, while smaller or larger $\beta_{max}$ both lead to higher Recs.

\subsubsection{Dimension of Code Book Vectors}
Finally, we vary latent dimension of $\{e_k\}_{k=1}^K$ in $\{8,16,32,64,128,256\}$, and the results are shown in Figure~\ref{fig:latent_dimension}. We find that the performance of language modelling is relatively robust to the choice of latent dimension. Intuitively, in the continuous space the dimension of latent variables is closely related to the model capacity. However, in the discrete case, the capacity of the model is largely determined by the code book size $K$ instead of the latent dimension, which is also verified in Table~\ref{table:density_results} and Figure~\ref{fig:two_stage}.

\section{Conclusion}
In this paper, we propose discrete variational attention model, a new algorithm for natural language generation. Our proposed approach can address the issues of  information underrepresentation and posterior collapse. Moreover we also carefully analyze the advantages of discreteness over continuity in variational attention models.
Extensive experiment results on benchmark language modelling datasets demonstrate the superiority of our proposed approach. As a future direction, our approach can be applied for more applications of natural language processing such as text summarization, dialogue systems and poetry generation.


\newpage
\bibliographystyle{named}
\bibliography{ijcai20}

\newpage
\end{document}